\documentclass[letterpaper]{article} % DO NOT CHANGE THIS
\usepackage{aaai2026}  % DO NOT CHANGE THIS
\usepackage{times}  % DO NOT CHANGE THIS
\usepackage{helvet}  % DO NOT CHANGE THIS
\usepackage{courier}  % DO NOT CHANGE THIS
\usepackage[hyphens]{url}  % DO NOT CHANGE THIS
\usepackage{graphicx} % DO NOT CHANGE THIS
\usepackage{natbib}  % DO NOT CHANGE THIS AND DO NOT ADD ANY OPTIONS TO IT
\usepackage{caption} % DO NOT CHANGE THIS AND DO NOT ADD ANY OPTIONS TO IT
\usepackage{algorithm}
\usepackage{algorithmic}

\usepackage{marvosym}
\usepackage{amsmath}
\usepackage{tabularx}
\usepackage{booktabs}
\usepackage{multirow}
\usepackage{amsfonts}
\newcolumntype{C}{>{\centering\arraybackslash}X}
\usepackage{newfloat}
\usepackage{listings}
\DeclareCaptionStyle{ruled}{labelfont=normalfont,labelsep=colon,strut=off} % DO NOT CHANGE THIS
\floatstyle{ruled}
\newfloat{listing}{tb}{lst}{}
\floatname{listing}{Listing}
\title{DeLightMono: Enhancing Self-Supervised Monocular Depth Estimation in Endoscopy by Decoupling Uneven Illumination}
\author{
    Mingyang Ou\textsuperscript{\rm 1}, 
    Haojin Li\textsuperscript{\rm 1}, 
    Yifeng Zhang\textsuperscript{\rm 1}, 
    Ke Niu\textsuperscript{\rm 2}, 
    Zhongxi Qiu\textsuperscript{\rm 3}, 
    Heng Li\textsuperscript{\rm 4,\Letter}, 
    Jiang Liu\textsuperscript{\rm 3,\rm 1}
}
\affiliations{
    \textsuperscript{\rm 1}Department of Computer Science and Engineering, Southern University of Science and Technology, Shenzhen, China\\
    \textsuperscript{\rm 2}Computer School, Beijing Information Science and Technology University, Beijing 100192, China\\
    \textsuperscript{\rm 3}Research Institute of Trustworthy Autonomous Systems, Southern University of Science and Technology, Shenzhen, China
    \textsuperscript{\rm 4}Faculty of Biomedical Engineering, Shenzhen University of Advanced Technology, Shenzhen, China\\
    \textsuperscript{\Letter} liheng@suat-sz.edu.cn
}

\usepackage{bibentry}
\begin{document}

\maketitle

\begin{abstract}
Self-supervised monocular depth estimation serves as a key task in the development of endoscopic navigation systems.
However, performance degradation persists due to uneven illumination inherent in endoscopic images, particularly in low-intensity regions.
Existing low-light enhancement techniques fail to effectively guide the depth network. Furthermore, solutions from other fields, like autonomous driving, require well-lit images, making them unsuitable and increasing data collection burdens.
To this end, we present DeLightMono - a novel self-supervised monocular depth estimation framework with illumination decoupling.
Specifically, endoscopic images are represented by a designed illumination-reflectance-depth model, and are decomposed with auxiliary networks.
Moreover, a self-supervised joint-optimizing framework with novel losses leveraging the decoupled components is proposed to mitigate the effects of uneven illumination on depth estimation.
The effectiveness of the proposed methods was rigorously verified through extensive comparisons and an ablation study performed on two public datasets.
\end{abstract}

% Uncomment the following to link to your code, datasets, an extended version or similar.
% You must keep this block between (not within) the abstract and the main body of the paper.
\begin{links}
    \link{Code}{https://github.com/ComgLq24/AAAI-2026-DeLightMono}
\end{links}

\section{Introduction}
Monocular endoscopy constitutes a fundamental component of disease diagnosis and intervention in minimally invasive surgery~\cite{zhu2021intelligent}.
Nevertheless, the confined anatomical structure of the human body greatly restricts the scope of view in monocular endoscopy and further increases the difficulty of intra-operative manipulation~\cite{yue2023deep}.
To alleviate the situation, surgical navigation~\cite{metzger2024augmented} and robotic-assisted surgery~\cite{lu2023autonomous} systems have been extensively researched to provide more perception and control under endoscopy.
While these systems ease the burden of surgeons, they both heavily rely on precisely estimating the depth.
\begin{figure}[!t]
    \centering
    \includegraphics[width=\linewidth]{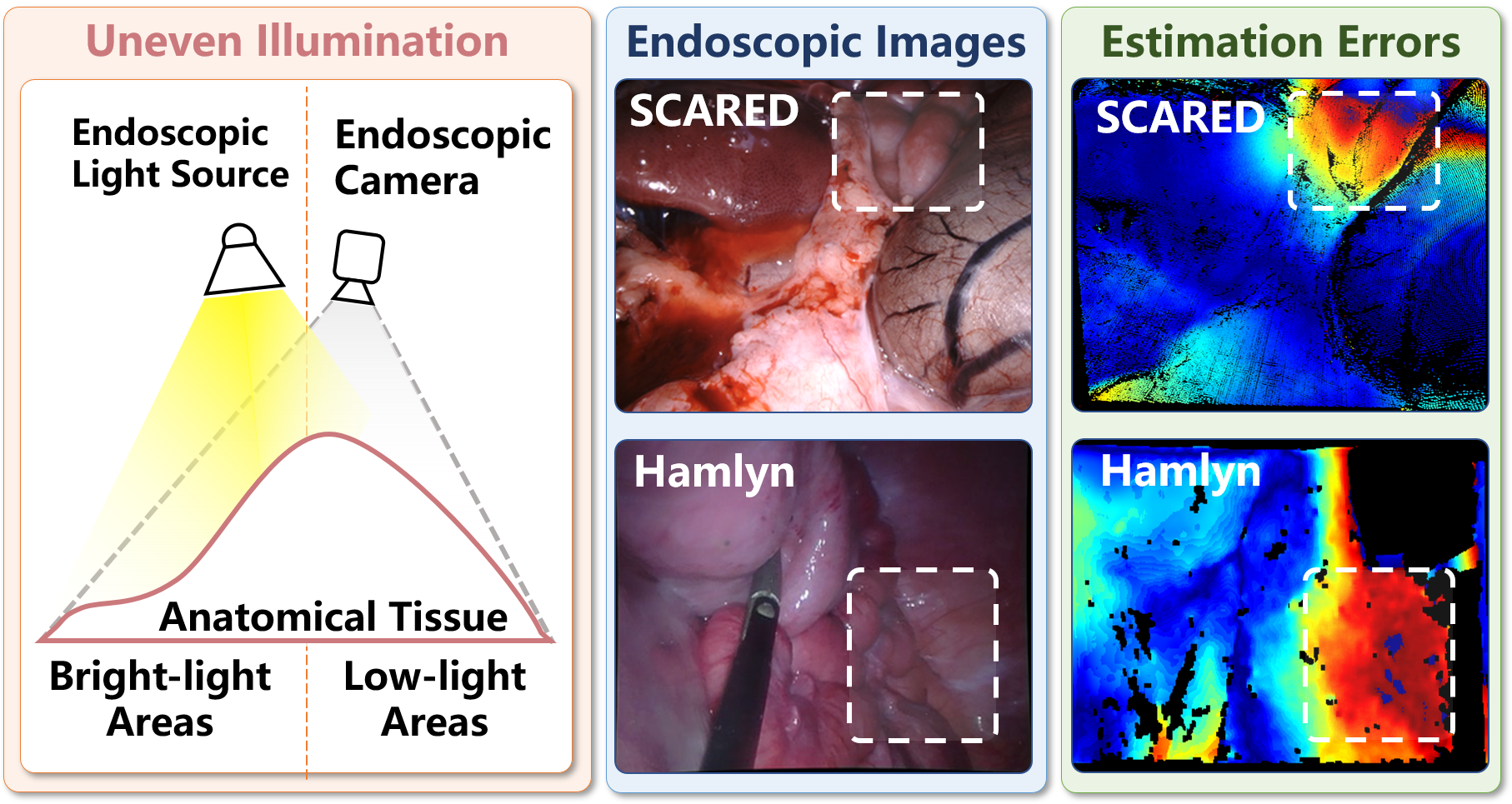}
    \caption{Uneven illumination affects depth estimation in endoscopic images. Left: illustration of uneven illumination. Middle: examples from SCARED and Hamlyn datasets. Right: error maps of Monodepth2, black areas contain nan depth label, where the highlighted regions suffer from elevated depth estimation errors.}
    \label{fig:challenge}
\end{figure}

Depth estimation from a sequence of frames has been a long-standing task in computer vision.
Traditional methods such as structure from motion(SfM)~\cite{enqvist2011non} and simultaneous localization and mapping(SLAM)~\cite{engel2014lsd} have been widely adopted for this purpose.
Recently, deep learning–based methods have shown remarkable potential in addressing the depth estimation task~\cite{fu2018deep,karsch2014depth}.
Pioneering works~\cite{li2015depth,eigen2014depth} train the depth net in a supervised manner.
However, in particular domains such as endoscopy, labeled data acquisition comes at a high cost, as the scene needs to be stationary while obtaining the ground truth.

Accordingly, self-supervised monocular depth estimation~\cite{zhou2017unsupervised} is proposed to mitigate the need for labeled data.
The depth network is trained by aligning 3D correspondences between the target frame and temporal adjacent source frames using depth cues.
Building upon this, a series of improvements~\cite{godard2019digging,lyu2021hr,amitai2023self} have been made by mining more informative geometric priors, scene-specific features to enhance depth estimation accuracy and generalization.
As for endoscopic images, researchers seek solutions to deal with challenges such as complex camera motion~\cite{li2022geometric}, specular reflection~\cite{li2024image}, textureless anatomy~\cite{he2024monolot}, etc.

However, these methods still struggle with precise depth estimation in endoscopic scenarios.
As illustrated in Fig.~\ref{fig:challenge}, monocular depth estimation suffers from uneven illumination in endoscopic images. 
Particularly, the estimation error increases in the low-light regions. 
Furthermore, the illumination level also varies across datasets, leading to a more challenging generalization problem.

While low-light enhancement methods~\cite{ma2022toward,wu2023learning} improve image visibility, offering a viable solution, they do not guide the depth net to learn the illumination distribution, resulting in suboptimal improvement.
Moreover, there exist methods~\cite{pmlr-v205-vankadari23a,zheng2023steps} facing a similar illumination problem in the natural scene; they require external resources such as bright light images or their corresponding depth maps during training, which is infeasible for endoscopy.

For these concerns, we propose a novel self-supervised framework(DeLightMono) with uneven illumination decoupling to enhance monocular depth estimation in endoscopic images. 
Firstly, we observe that endoscopic images are predominantly illuminated by a single artificial light source.
The insight drives our Illumination-Reflectance-Depth (IRD) model, which decouples the illumination-invariant reflectance network from the image.
We then introduce a unified self-supervised framework for jointly optimizing the IRD decomposition and monocular depth estimation.
Specifically, the IRD decoupling branch involves a reconstruction loss and an intensity ratio consistency loss with two auxiliary networks to correctly estimate illumination and reflectance.
On the other hand, the depth estimation branch benefits from a redesigned photometric loss and an illumination-map guided degradation consistency loss, both aimed at mitigating uneven illumination. Our major contributions are as follows:
\begin{itemize}
    \item To alleviate the negative impact of uneven illumination on monocular depth estimation, we design an illumination-reflectance-depth(IRD) modeling for endoscopic images.
    \item Building on this modeling, a jointly optimized self-supervised framework is proposed for depth estimation and IRD decoupling, where the decoupled IRD components guide the depth net to correctly estimate the depth from uneven illumination by four novel losses.
    \item The robustness of the proposed framework against uneven illumination is verified by a comprehensive comparison with state-of-the-art methods and an ablation study on two public endoscopic depth datasets.
\end{itemize}

\section{Related Work}
\subsection{Self-Supervised Monocular Depth Estimation}

Predicting 3D depth information from 2D images has been a fascinating topic in 3D vision. 
Early-stage methods~\cite{eigen2014depth,li2015depth,karsch2014depth} leveraged annotated datasets to supervise the training of the depth estimation model.

As obtaining labeled depth data is expensive, \citet{godard2017unsupervised} took the first step toward alleviating this issue by enforcing the left-right consistency across unlabeled stereo images.
\citet{zhou2017unsupervised} extended the idea to monocular settings by employing an auxiliary camera pose estimation network.
Subsequent works focus on fusing higher-resolution features~\cite{lyu2021hr,zhao2022monovit} and developing lightweight architectures for edge devices~\cite{zhao2022monovit}.
In addition,  various cues such as dynamic and static objects~\cite{klingner2020self,godard2019digging,zhou2025manydepth2} and depth uncertainty~\cite{poggi2020uncertainty} were incorporated in the training process to improve overall performance.

Unlike natural images, an endoscopic scene poses greater challenges for depth estimation.
Some methods introduced keypoint matching losses~\cite{he2024monolot} or fused global-local features~\cite{fan2024triple} to address the distortion of low-texture regions.
Others employed the Siamese network~\cite{li2022geometric} and transformation consistency module~\cite{yue2023tcl} to learn a more accurate camera pose.
As varying specular reflection breaks the photometric consistency assumption, researchers adopted appearance flow~\cite{shao2022self,ozyoruk2021endoslam} networks and retinex theory~\cite{li2024image} to mitigate the effects.
In recent years, several studies have also explored transferring depth estimation foundation models~\cite{yang2024depth,ranftl2020towards} to endoscopic datasets~\cite{cui2024endodac,budd2024transferring}.

However, existing methods rarely consider the negative impact of spatially uneven illumination in each frame(as in Fig.~\ref{fig:challenge}), which prohibits the depth net from learning the correct depth in the low-light regions. Moreover, the appearance of low-light regions remains stable across adjacent frames, yet depth accuracy drops significantly in these regions. Consequently, prevalent methods that study varying specular reflection are insufficient to address this issue.

\subsection{Low-Light Image Enhancement}
The objective of low-light image enhancement is to improve the image quality by recovering visual details in the dark regions.
Traditional methods like CLAHE~\cite{pizer1987adaptive,reza2004realization} and Lime~\cite{guo2016lime} calculated the statistical information or optimized an illumination map with a mathematical model to redistribute the pixel intensity.
However, this type of method can amplify noise in images, especially in uniform or low-contrast regions, leading to a degradation in overall image quality. Additionally, their effectiveness is highly sensitive to parameters. 
\begin{figure*}[!ht]
    \centering
    \includegraphics[width=0.90\linewidth]{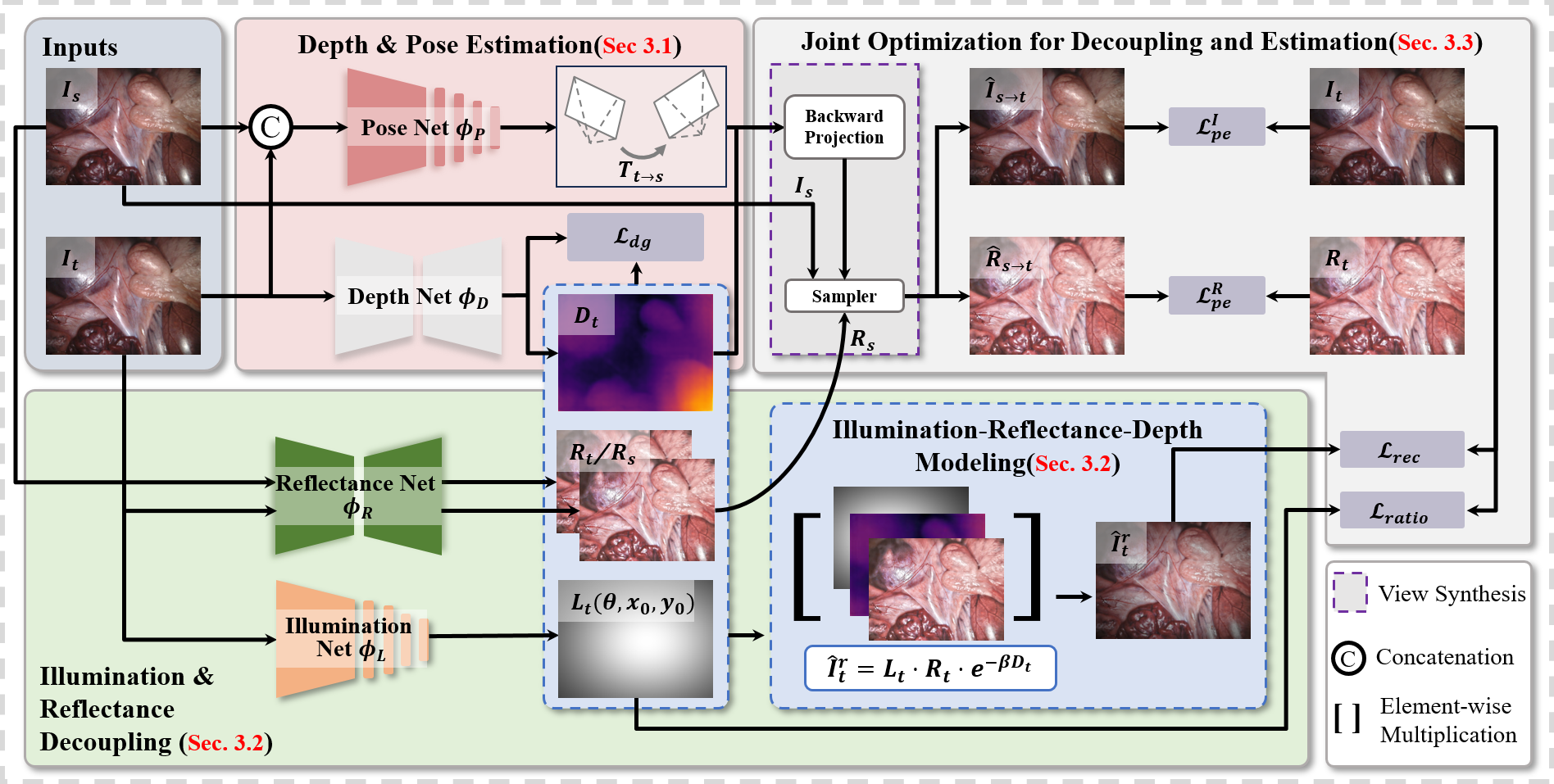}
    \caption{Overview of the proposed method.}
    \label{fig:main}
\end{figure*}
Alternative data-driven approaches incorporated the Retinex theory to decouple illumination and reflectance maps, thereby enabling effective illumination correction~\cite{zhang2021beyond,Chen2018Retinex,cai2023retinexformer}.
These methods all require a paired normal-/low-light images dataset for supervised training, whereas most recently, \citet{ma2022toward} and \citet{guo2020zero} developed unsupervised methods and iteratively lit-up the low-light images.

\section{Method}
This section first reviews the conventional approach to self-supervised monocular depth estimation (Sec.~\ref{sec:conv_method}).
Noting its poor performance in endoscopy due to uneven illumination, we introduce Illumination-Reflectance-Depth (IRD) modeling to explicitly connect illumination, reflectance, and depth.
Based on this model, we design an auxiliary branch with two networks to decouple the illumination and reflectance maps from an endoscopic image (Sec.~\ref {sec:modeling}).
Finally, we propose a joint self-supervised framework that employs novel loss functions to mutually optimize the IRD decoupling branch and the depth estimation branch (Sec.~\ref{sec:joint_model}).
\subsection{Self-Supervised Monocular Depth Estimation}
\label{sec:conv_method}
Following~\cite{godard2019digging}, self-supervised monocular depth estimation learns the depth from the reconstruction loss between the target frame and a synthesized view from the adjacent frames.
As shown in the red block of Fig.~\ref{fig:main}, given a target endoscopic frame $I_t$ and its adjacent frames $I_s\in \{t+1,~t-1\}$ in a continuous sequence, the depth map is estimated by a depth net $\phi_D$. Meanwhile, the camera pose transformation $T_{t\rightarrow s}$ between $I_t$ and $I_s$ is also predicted by a pose net $\phi_P$. Subsequently, the pixel-wise correspondence regarding the target frame $I_t$ and the source frame $I_s$ can be constructed via the following equation:
\begin{equation}
    \label{eq:pixel_corr}
    p_s\sim{KT_{t\rightarrow{s}}}{D_t}(p_t){K^{-1}}p_t,
\end{equation}

\noindent
where $K$ is the camera intrinsic, $p_t$/$p_s$ are corresponding pixels in $I_t/I_s$ respectively. 
Based on Eq.~\ref{eq:pixel_corr}, a reconstructed target frame $\hat{I}_{s\rightarrow t}$ is synthesized by bilinear sampling from adjacent source frames $I_s$. 

As inaccurate depth will result in dissimilarity between $I_t$ and $\hat{I}_{s\rightarrow t}$, a photometric reconstruction loss can be calculated to optimize the depth net $\phi_D$ and the pose net $\phi_P$:
\begin{equation}
    \label{eq:pei}
    \mathcal{L}_{pe}^I = \gamma(1-SSIM(\hat{I}_{s\rightarrow t},~I_t)+(1-\gamma)||\hat{I}_{s\rightarrow t}-I_t||_1),
\end{equation}

\noindent
where $SSIM$ measures the structure similarity between $\hat{I}_{s\rightarrow t}$ and $I_t$ and $\gamma$ sets to 0.85 following existing methods~\cite{godard2019digging,shao2022self}.

Moreover, an edge-aware smoothness loss is also adopted to encourage the sharpness of depth on the boundary of the object:
\begin{equation}
    \label{eq:edge}
    \mathcal{L}_{edge} = |\partial_xD_t|e^{-|\partial_xI_t|}+|\partial_yD_t|e^{-|\partial_yI_t|}.
\end{equation}

\subsection{Illumination-Reflectance-Depth Modeling for Endoscopic Images}
\label{sec:modeling}

Although this standard paradigm yields promising results in the natural image domain, its performance degrades in endoscopic scenes due to the uneven illumination.

While low-light enhancement can improve visibility~\cite{reza2004realization,ma2022toward}, they do not provide explicit guidance for learning illumination distributions. 
Moreover, approaches from other domains that handle lighting variations~\cite{wang2021regularizing, liu2021self} rely on image collections captured under diverse lighting conditions, which are costly to acquire for endoscopic scenes. 
These limitations motivate our design of an illumination model tailored to endoscopy, aiming to improve depth estimation under uneven illumination without requiring specialized data.

\subsubsection{Illumination-Reflectance-Depth (IRD) Model:}
According to the Retinex theory~\cite{land1971lightness,land1977retinex}, an image can be represented by the element-wise Hadamard product of a reflectance map and an illumination map:
\begin{equation}
    \label{eq:retinex}
    \hat{I}^r_t = R_t \cdot L_t,
\end{equation}
\noindent
where $R$ is the reflectance map and $L$ is the illumination map. 
Furthermore, endoscopic images are a type of near-field imaging and are often collected in a confined space.
Consequently, anatomical structures farther from the camera tend to receive fewer light rays.
Based on this observation, we consider that depth information also plays a key role in the appearance of the images.
Then, we can extend the Retienx theory~\cite{land1971lightness,land1977retinex} to the following formulation for the endoscopic images:
\begin{equation}
    \label{eq:retinex_D}
    \hat{I}^r_t = R_t\cdot L_t\cdot e^{-\beta D_t},
\end{equation}
\noindent
where $\beta$ is the depth scaling factor that prevents $D_t$ from over-dominant.

Another prominent characteristic of endoscopic images is their exclusive reliance on artificial illumination, with no contribution from ambient light sources. 
In the Lambertian model~\cite{oren1995generalization}, the single source light can be formulated by a cosine function with an angular decay factor:
\begin{equation}
    \label{eq:L}
    L_t = cos^\theta(I_t, L^c_t) = (\frac{1}{1+||I_t-L^c_t)||_2})^\theta,
\end{equation}
\noindent
where $L_t^c$ is the center of the artificial light source, $\theta$ is the angular decay factor that controls the degradation speed of illumination.

\subsubsection{IRD Decoupling:}
As can be seen in the green block of Fig.~\ref {fig:main}, we set two auxiliary networks $\phi_L$ and $\phi_R$ to predict the illumination and reflectance map.

\begin{equation}
    \label{eq:decouple}
    \begin{aligned}
        R_t &= \phi_R(I_t)\\
        (\theta, x_0, y_0) &= \phi_L(I_t).
    \end{aligned}
\end{equation}
\noindent Notably, the illumination net estimates the parameters of the illumination map, and $(x_0, y_0)$ are the pixel coordinates of the center $L_t^c$.
Combined with the depth map, we can decouple the endoscopic images from the IRD modeling and recover the texture details of low-light regions in the endoscopic images.

\subsection{Joint Optimization of IRD Decoupling and Depth Estimation}
\label{sec:joint_model}
Based on the IRD model, we propose a self-supervised joint optimization framework for monocular depth estimation and IRD decoupling.
\begin{equation}
    \label{eq:joint}
    \begin{cases}
        \hat{I}_t^r &=R_t \cdot L_t \cdot e^{-\beta D_t} \\
        R_t&\stackrel{\text{sampling}}{\leftarrow}(R_s, p_s),\,\,p_s\sim{KT_{t\rightarrow{s}}}{D_t}(p_t){K^{-1}}p_t
    \end{cases}
\end{equation}

As shown in formula Eq.~\ref{eq:joint}, the benefits are twofold.
On one hand, as the depth map is part of the IRD model, the more precise the depth prediction is, the better the reflectance and illumination map can be learned.
On the other hand, since the decoupled reflectance recovers the intrinsic appearance in low-light regions, it can be used for view synthesis between adjacent frames and target frames.
By calculating photometric loss on the synthesized reflectance and the true reflectance, the depth net and the pose net receive the correct gradient backpropagation from the low-light regions, thus having better performance.
Additionally, the illumination map can be applied to the original RGB images, serving as a simple yet powerful augmentation to strengthen the generalized ability of the depth net.

\subsubsection{IRD decoupling branch: } We propose two losses for training the IRD decoupling branch.
With each component of the IRD model estimated by the depth net and two auxiliary networks, the reconstructed target frame $\hat{I}_t^r$ is acquired by element-wise multiplication.
Subsequently, a reconstruction loss is calculated between the $\hat{I}_t^r$ and the real target frame $I_t$:
\begin{equation}
    \label{eq:loss_rec}
    \begin{aligned}
        \hat{I}^r_t &= \phi_R(I_t) \cdot L_t^c(\phi_L(I_t))\cdot e^{\beta D_t}\\
        \mathcal{L}_{rec}&= \gamma(1-SSIM(\hat{I}^r,~I_t)+(1-\gamma)||\hat{I}^r-I_t||_1)
    \end{aligned}
\end{equation}
where $L_t^c(\cdot)$ builds the illumination map from the estimated $(x_0, y_0)$, and $\theta$ parameters. 

However, simply relying on reconstruction loss may easily fall into the trivial solution, where all predicted illumination maps remain consistent and the reflectances are the same as the target frame $I_t$. 
As a solution, the illumination net is regularized by comparing the intensity ratio between the target frame $I_t$ and the illumination map $L_t$. 
To be more concrete, with $p_{95}$ and $p_{20}$ being the percentile intensity value, we hypothesize that a correct illumination map should have a ratio between high and low intensity similar to that of the target frame $I_t$.
\begin{equation}
    \label{eq:loss_int_1}
    \mathcal{L}_{ratio} = ||\frac{p_{95}(I_t)}{p_{20}(I_t)}-\frac{p_{95}(L_t)}{p_{20}(L_t)} ||_1
\end{equation}

By these two loss functions, we can prevent the trivial solution and decouple well-illuminated reflectance in a self-supervised manner.

\subsubsection{Depth estimation branch: } 
Illustrated in the gray block of Fig.~\ref {fig:main}, we consider how the decoupled IRD component can be used to enhance the performance of the depth estimation model.
As mentioned previously, we notice that the reflectance recovers the texture details in the dark areas.
Thereafter, we can synthesize target reflectance $\hat{R}_{s\rightarrow t}$ from the source reflectance $R_s$, and calculate the photometric loss similar to Eq.~\ref{eq:pei}.
\begin{equation}
    \label{eq:per}
    \mathcal{L}_{pe}^R = \gamma(1-SSIM(\hat{R}_{s\rightarrow t},~R_t)+(1-\gamma)||\hat{R}_{s\rightarrow t}-R_t||_1),
\end{equation}

By adding this auxiliary loss, the depth and pose net will receive correct backpropagation gradients in the low-light regions.
In addition, to further enhance the robustness of the depth estimation model against uneven illumination, we directly combine the predicted depth $D_t$ and illumination maps $L_t$ with the original target frame to generate a degraded image $\hat{I}_t^{dg}$.
\begin{equation}
    \label{eq:degrade}
        \hat{I}_t^{dg} =I_t \cdot L_t \cdot e^{-\beta D_t} \\
\end{equation}

We hypothesize that if the depth net is robust enough to the uneven illumination, the depth output from $I_t$ and $\hat{I}^{dg}_t$ should be the same. 
Therefore, a degradation consistency loss is introduced:
\begin{equation}
    \label{eq:loss_int_2}
    \mathcal{L}_{dg} = ||\phi_D(I_t)-\phi_D(\hat{I}_t^{dg})||_1
\end{equation}

\subsubsection{Overall Loss:} Finally, we obtain the overall losses for networks of both the depth estimation branch and IRD decoupling branch.
\begin{equation}
    \label{eq:overall}
    \begin{cases}
        \mathcal{L}_{\phi_D} = \mathcal{L}_{pe}^I+\mathcal{L}_{pe}^R+\mathcal{L}_{dg} \\
        \mathcal{L}_{\phi_P} = \mathcal{L}_{pe}^I+\mathcal{L}_{pe}^R \\
        \mathcal{L}_{\phi_R} = \mathcal{L}_{rec}\\
        \mathcal{L}_{\phi_L} = \mathcal{L}_{rec}+\mathcal{L}_{ratio}
    \end{cases}
\end{equation}

\section{Experiments}
\subsection{Datasets}
We evaluate our method using two public endoscopic datasets for both comparative and ablative studies.

\textbf{SCARED:} The SCARED~\cite{allan2021stereo} dataset consists of stereo laparoscopic videos acquired from fresh porcine cadaver abdominal anatomy, with 35 stereo videos recorded across 9 distinct scenes. Adhering to previous work~\cite{huang2022self}, we select 15351 frames for training, 1705 for validation, and 90 keyframes with accurate depth labels acquired using structured light for testing.

\textbf{Hamlyn:} The Hamlyn~\cite{recasens2021endo} dataset contains in vivo stereo endoscopic videos captured during various surgical procedures. It poses greater challenges due to low-light conditions and significant inter-frame motion. In line with prior protocols~\cite{cui2024endodac,recasens2021endo}, we use all 21 videos for generalization evaluation.

\subsection{Implementations}
We implement our method with Pytorch on a single NVIDIA RTX A6000. 
Adam~\cite{kingma2020method} optimizer is used with $\beta_1$=0.9, $\beta_2$=0.99.
The initial learning rate is set to 1e-4 and scales down by 0.3 every 10 epochs. 
The total training epoch is 30 with a batch size of 12.
Cohere to previous methods~\cite{godard2019digging,cui2024endodac}, preprocessing contains resizing the image to 320$\times$256, randomly flipping, and color jittering.
We select two different base models for a comprehensive evaluation: Monodepth2~\cite{godard2019digging} with ResNet-18, and DepthAnything~\cite{yang2024depth} with Vit-B.
In addition, the implementation of illumination net $\phi_L$ and the pose net $\phi_P$ is a simple ResNet-18 regression network, while the reflectance network $\phi_R$ is adopted from the RetinexNet~\cite{Chen2018Retinex}.

\subsection{Metrics}
We adopt commonly-used depth estimation indicators, including Abs Rel, Sq Rel, RMSE, RMSE log errors, and the $\delta<1.25^1$, $\delta<1.25^2$, $\delta<1.25^3$ accuracy metrics, to comprehensively assess model performance. See the supplement for more details.

\subsection{Effectiveness of Illumination Decoupling}
While low-light enhancement methods from natural image processing offer potential for endoscopic illumination, our analysis shows their limitations and highlights the need for accurate illumination modeling in endoscopic scenes.
\begin{table}[!ht]
  \centering
  \fontsize{9pt}{9pt}\selectfont
  \setlength{\tabcolsep}{1mm}
  \begin{tabularx}{1.0\linewidth}{@{}l@{\;}|cc|cc@{}}
    \hline
    Methods&Abs Rel$\downarrow$&Sq Rel$\downarrow$&$\delta<1.25^1$$\uparrow$&$\delta<1.25^2$$\uparrow$ \\
    \hline
    Monodepth2(-)&0.075&0.799&0.952&0.990\\
    CLAHE(\textbf{S})&0.072&0.787&0.958&0.991\\
    LIME(\textbf{S})&0.077&0.819&0.949&0.992\\
    SCI(\textbf{D})&0.073&0.790&0.954&0.992\\
    DeLightMono(\textbf{D})&\textbf{0.065}&\textbf{0.655}&\textbf{0.961}&\textbf{0.993}\\
    \hline
    \multicolumn{5}{@{}p{0.97\linewidth}}{\footnotesize \textbf{S}: statistics-based method, \textbf{D}: learning-based method. $\uparrow$: the higher the better, $\downarrow$: the lower the better. The best results are reported in \textbf{bold}.}\\
  \end{tabularx}
  \caption{Performance comparison with image enhancement methods. All models are based on Monodepth2 and evaluated on the SCARED dataset.}
  \label{tab:enh}
\end{table}
\begin{figure}[!ht]
    \centering
    \includegraphics[width=0.99\linewidth]{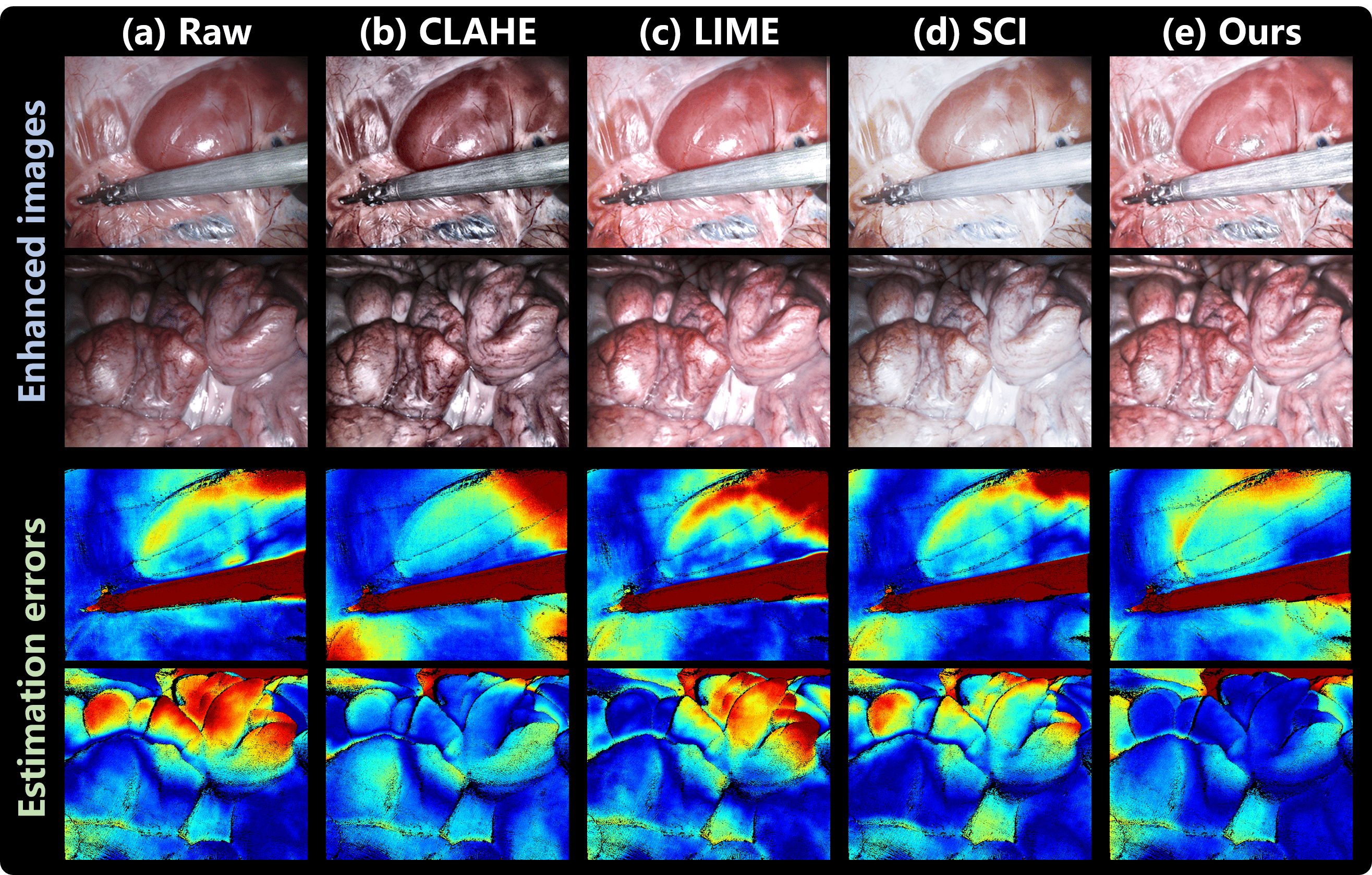}
    \caption{Enhancement results and error maps on the SCARED dataset.}
    \label{fig:ablation_enhanced}
\end{figure}

Two traditional statistics-based methods(CLAHE~\cite{reza2004realization} and LIME~\cite{guo2016lime}) and one unsupervised learning-based method(SCI~\cite{ma2022toward}) are selected for comparison. 
All these methods serve as a preprocessing strategy during training and inference.
As Tab.~\ref{tab:enh} demonstrates, compared to the baseline Monodepth2~\cite{godard2019digging}, the statistic-based method CLAHE~\cite{reza2004realization} and the learning-based method SCI~\cite{ma2022toward} both slightly improve the performance.
However, such amelioration is minor, and for LIME~\cite{guo2016lime}, the situation is even deteriorating.

A more intuitive illustration is shown in Fig.~\ref{fig:ablation_enhanced} with enhancement results and error maps.
From the top cases, CLAHE~\cite{reza2004realization} amplified the texture details in the image, while SCI~\cite{ma2022toward} and LIME~\cite{guo2016lime} lit up the illumination level globally.
However, row 1 shows that neither the light level nor the texture details are recovered in the dark areas of these methods' results, such as the top-right corner.
Additionally, SCI~\cite{ma2022toward}, trained on the SCARED~\cite{allan2021stereo} before being applied, introduces color distortion.
Consequently, estimation error does not decline and even increases (row 3, columns b and d).
\begin{table*}[!ht]
  \fontsize{9pt}{9pt}\selectfont
  \centering
  \begin{tabularx}{1.0\textwidth}{l|@{\ \;}c@{\ \;}|@{\ \;}c@{\ \;\ }|c@{\ \;\ }c@{\ \;\ }c@{\ \;}c|ccc}
    \hline
    Method&Year&Backbone&Abs Rel$\downarrow$&Sq Rel$\downarrow$&RMSE$\downarrow$&RMSE log$\downarrow$&$\delta<1.25^1$$\uparrow$&$\delta<1.25^2$$\uparrow$&$\delta<1.25^3$$\uparrow$\\
    \hline
    Monodepth2&2019&ResNet-18&0.075&0.794&6.213&0.099&0.952&0.991&0.998\\
    HR-Depth&2021&ResNet-18&0.069&0.731&5.777&0.093&0.959&0.993&0.997\\
    M3Depth&2022&ResNet-18&0.072&0.764&6.178&0.098&0.961&0.992&0.996\\
    AF-SfMLearner&2022&ResNet-18&0.075&0.798 &6.212&0.100&0.953&0.991&0.997\\
    IID-SfMLearner&2024&ResNet-18&0.072& 0.789& 5.938& 0.096& 0.961& 0.991&0.996\\
    \hline
    DeLightMono(Ours)&-&ResNet-18&\textbf{0.065}&\textbf{0.655}&\textbf{5.589}&\textbf{0.089}&\textbf{0.961}&\textbf{0.993}&\textbf{0.998}\\ 
    \hline
    DA(zero-shot)&2024&ViT-B&0.197&4.685&14.60&0.245&0.729&0.903&0.965\\
    DA(finetune)&2024&ViT-B&0.065&0.563&5.640&0.088&0.962&\textbf{0.997}&0.999\\
    MonoViT&2022&ViT-B&0.065&0.642&5.492&0.088&0.961&0.994&0.998\\
    EndoDAC&2024&ViT-B&0.064& 0.607& 5.715& 0.088& 0.963& 0.995&0.998\\
    \hline
    DeLightMono(Ours)&-&ViT-B&\textbf{0.059}&\textbf{0.496}&\textbf{5.214}&\textbf{0.081}&\textbf{0.969}&\textbf{0.997}&\textbf{1.000}\\
    \hline
    \multicolumn{10}{p{0.97\linewidth}}{\footnotesize $\uparrow$: the higher the better, $\downarrow$: the lower the better. The best results are reported in \textbf{bold}.}
  \end{tabularx}
  \caption{Quantitative results on the SCARED dataset.}
  \label{tab:quan_scared}
\end{table*}
\begin{figure*}[!ht]
    \centering
    \includegraphics[width=0.92\linewidth]{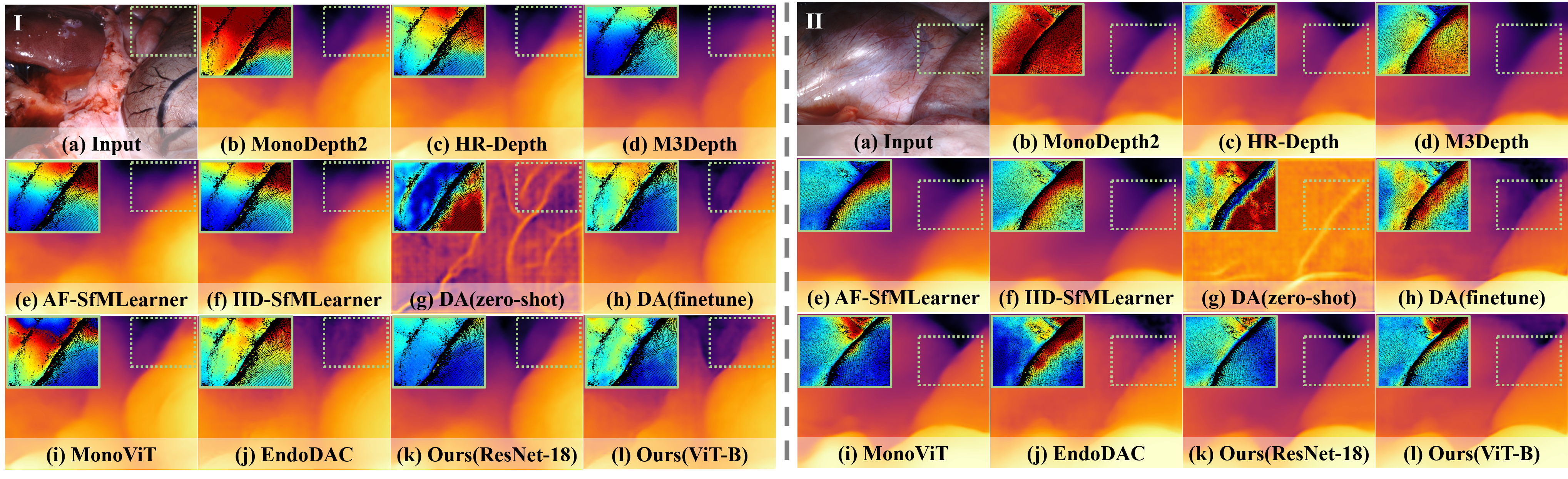}
    \caption{Qualitative comparisons on the SCARED dataset with error maps of the highlighted regions.}
    \label{fig:qualitative_scared}
\end{figure*}
For another case, as in rows 2 and 4, LIME~\cite{guo2016lime} successfully recovers the light level and details at the top of the image.
Nevertheless, the performance improvement is also limited.

These findings imply the necessity of modeling the illumination and utilizing the illumination as a guide during the training of the depth estimation network.
In contrast, our proposed framework jointly optimizes the IRD decoupling and depth estimation task, resulting in restored details and performance enhancement in dark areas, as evidenced in rows 3 and 4 of Fig.~\ref{fig:ablation_enhanced}.

\subsection{Compare with SOTA methods}
\subsubsection{Comparison on SCARED dataset:} 
Tab.~\ref{tab:quan_scared} shows a comparison with the state-of-the-art methods both from the natural-image and endoscopic domains on the SCARED~\cite{allan2021stereo} dataset.
Monodepth2~\cite{godard2019digging}, as the baseline method, suffers from performance degradation due to the uneven illumination, while HR-Depth~\cite{lyu2021hr} redesigns the skip-connection to fuse high-resolution features, thereby achieving better results.
M3depth~\cite{huang2022self} focuses on 3D geometric consistency between frames, while Af-SfMLearner~\cite{shao2022self} and IID-SfMLearner~\cite{li2024image} are solutions for varying specular reflection between adjacent frames.
As they do not account for the spatially uneven illumination, their improvements are also marginal.
The depth foundation model Depth anything~\cite{yang2024depth} exhibits robust zero-shot ability across various scenes.
Nevertheless, its performance declined significantly, which shows the gaps between natural images and endoscopic images.
After finetuning, it shows similar performance compared to MonoVit~\cite{zhao2022monovit} and EndoDAC~\cite{cui2024endodac}.

Our DeLightMono(ViT-B) achieves the best performance across all metrics.
What is more noteworthy is that our DeLigthMono(ResNet-18) achieves comparative results with ViT-B-based methods, which validates the feasibility of our illumination model and the joint optimization framework.

Furthermore, we visualize the depth outputs and their error maps in low-light regions in Fig.~\ref{fig:qualitative_scared}. 
As the angle between the regions and the light source increases, their local illumination diminishes significantly, leading to a textureless appearance.
Since other methods do not take the varying illumination of endoscopic images into account, their error rates grow correspondingly.
On the contrary, benefitting from the reflectance and our joint optimization framework, our methods not only maintain superior results but also have clearer depth prediction in these regions.

\subsubsection{Comparison on Hamlyn dataset:} 
\begin{table*}[!t]
  \fontsize{9pt}{9pt}\selectfont
  \centering
  \begin{tabularx}{1.0\textwidth}{l|@{\ \;}c@{\ \;}|@{\ \;}c@{\ \;\ }|c@{\ \;\ }c@{\ \;\ }c@{\ \;}c|ccc}
    \hline
    Method&Year&Backbone&Abs Rel$\downarrow$&Sq Rel$\downarrow$&RMSE$\downarrow$&RMSE log$\downarrow$&$\delta<1.25^1$$\uparrow$&$\delta<1.25^2$$\uparrow$&$\delta<1.25^3$$\uparrow$\\
    \hline
    Monodepth2&2019&ResNet-18&0.165&4.586&16.973&0.218&0.763&0.919&0.974\\
    HR-Depth&2021&ResNet-18&0.162&4.416&16.713&0.214&0.767&0.922&0.976\\
    M3Depth&2022&ResNet-18&0.161&4.377&16.497&0.213&0.768&0.921&0.976\\
    AF-SfMLearner&2022&ResNet-18&0.179&5.665&17.693&0.226&0.759&0.913&0.968\\
    IID-SfMLearner&2024&ResNet-18&0.168&4.787&16.944&0.217&0.767&0.921&0.974\\
    \hline
    DeLightMono(Ours)&-&ResNet-18&\textbf{0.158}&\textbf{4.095}&\textbf{16.127}&\textbf{0.208}&\textbf{0.773}&\textbf{0.929}&\textbf{0.980}\\
    \hline
    DA(zero-shot)&2024&ViT-B&0.174&4.473&18.307&0.238&0.718&0.912&0.975\\
    DA(finetune)&2024&ViT-B&0.146&\textbf{3.546}&15.217&0.198&0.789&0.936&\textbf{0.984}\\
    MonoViT&2022&ViT-B&0.158&4.215&16.403&0.211&0.776&0.927&0.978\\
    EndoDAC&2024&ViT-B&0.146&3.861&15.058&\textbf{0.194}&\textbf{0.802}&\textbf{0.937}&0.980\\
    \hline
    DeLightMono(Ours)&-&ViT-B&\textbf{0.144}&3.648&\textbf{14.989}&0.196&0.799&\textbf{0.937}&0.981\\
    \hline
    \multicolumn{10}{p{0.97\linewidth}}{\footnotesize $\uparrow$: the higher the better, $\downarrow$: the lower the better. The best results are reported in \textbf{bold}.}
  \end{tabularx}
  \caption{Generalization results on the Hamlyn dataset. All methods are trained on the SCARED dataset.}
  \label{tab:quan_hamlyn}
\end{table*}
\begin{figure*}[!ht]
    \centering
    \includegraphics[width=0.88\linewidth]{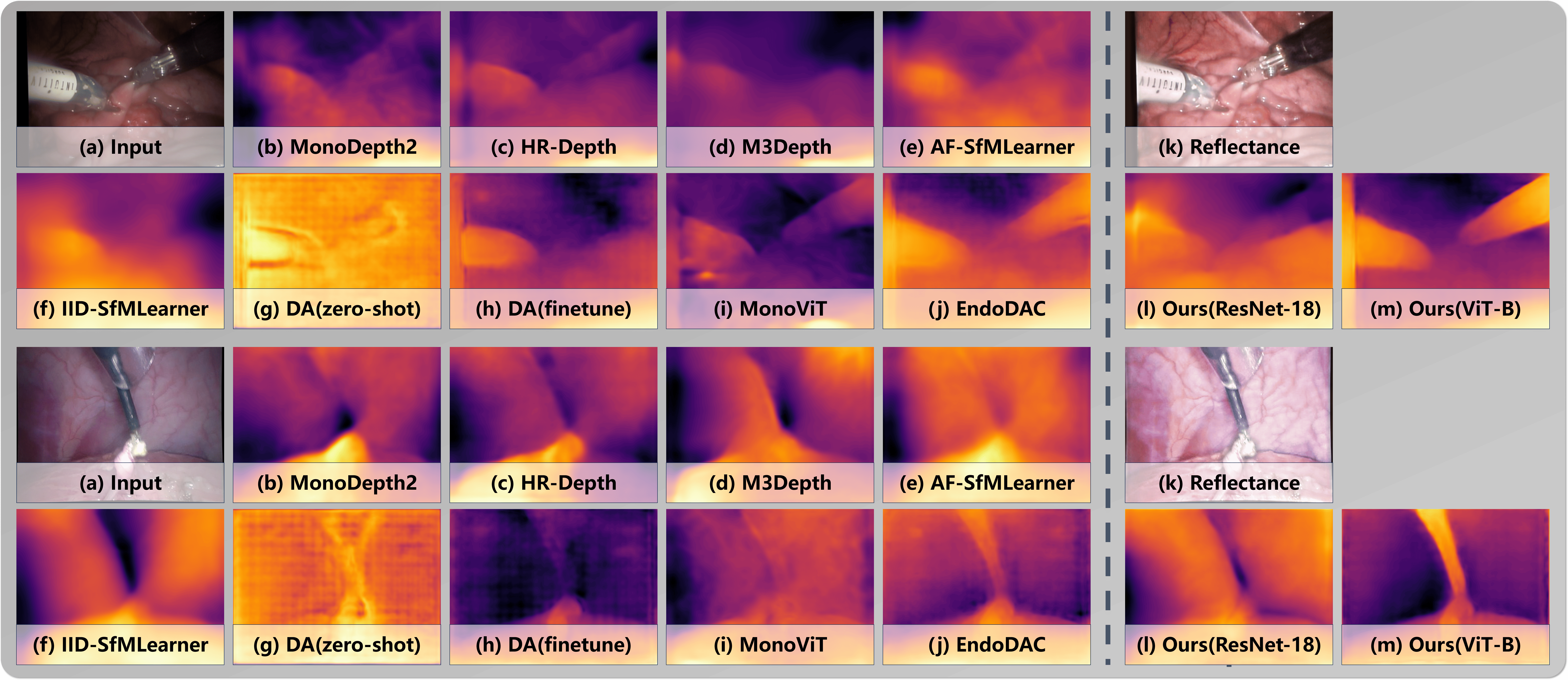}
    \caption{Qualitative comparisons on the Hamlyn dataset. Additionally, reflectances from the proposed method are provided.}
    \label{fig:qualitative_hamlyn}
\end{figure*}
Fig.~\ref{fig:qualitative_hamlyn} and Tab.~\ref{tab:quan_hamlyn} shows the generalization comparisons on the Hamlyn~\cite{recasens2021endo} dataset.
As observed, Hamlyn~\cite{recasens2021endo} presents a significant difference in appearance compared to SCARED~\cite{allan2021stereo}, especially the illumination levels.
Comparison models with the ResNet-18 backbone struggle to recognize the shapes of tissues and surgical instruments in the depth maps.
Conversely, our model still roughly captures edges and the relative depth.
Benefiting from contextual understanding of the vision transformer, ViT-B-based models offer substantial alleviation of the situation.
Combined with our proposed method, the depth net accurately delineates instrument shapes and makes precise depth predictions. 
Furthermore, the quantitative results presented in the data table align well with the visualized outputs, and our models achieve leading performance across most of the evaluation metrics.
\subsection{Ablation studies}
To validate the contribution of each component to our proposed methods, ablation studies are conducted on SCARED~\cite{allan2021stereo} by removing each of them in turn.
The results are reported in Tab.~\ref{tab:ablation}.

Comparing rows 2 and 6, the performance drops if we remove the reflectance-based photometric loss term.
The result shows that the recovered illumination and texture in low-light regions act as key factors in enhancing the performance of the depth estimation model.
Meanwhile, a similar performance deterioration takes place when comparing rows 3 and 6, further demonstrating that our degradation consistency loss contributes to enhancing the model's robustness against illumination variations in endoscopic images.

We observe a more drastic decline in estimation accuracy once the $\mathcal{L}_{ratio}$ or $\mathcal{L}_{rec}$ is removed. 
More specifically, the illumination map becomes a uniform distribution on the model w/o $L_{ratio}$, while the reflectance loses most of the semantic information if the $\mathcal{L}_{rec}$ is left.
This indicates that both $\mathcal{L}_{ratio}$ and $\mathcal{L}_{rec}$ serve as the key regularization terms for proper IRD decoupling.
\begin{table}[!ht]
    \fontsize{9pt}{9pt}\selectfont
    \centering
    \begin{tabular}{@{\,}c@{\;}c@{\;}c|c@{\;}c||c@{\quad}c@{\quad}c@{\,}}
    \hline
    \multicolumn{3}{c|}{Depth} & \multicolumn{2}{c||}{Decoupling}&\multirow{2}{*}{Abs Rel$\downarrow$}&\multirow{2}{*}{Sq Rel$\downarrow$}&\multirow{2}{*}{RMSE$\downarrow$}\\
    \cline{1-5}
    $L^I_{pe}$&$L^R_{pe}$&$L_{dg}$&$L_{rec}$&$L_{ratio}$&&&\\
    \hline
    \checkmark&&&&&0.075&0.794&6.213\\
    \hline
    \checkmark&&\checkmark&\checkmark&\checkmark&0.067&0.682&5.631\\
    \checkmark&\checkmark&&\checkmark&\checkmark&0.067&0.692&5.629\\
    \checkmark&\checkmark&\checkmark&&\checkmark&0.069&0.712&5.812\\
    \checkmark&\checkmark&\checkmark&\checkmark&&0.069&0.744&5.860\\
    \hline
    \checkmark&\checkmark&\checkmark&\checkmark&\checkmark&\textbf{0.065}&\textbf{0.655}&\textbf{5.589}\\
    \hline
    \end{tabular}
    \caption{Ablations on the SCARED dataset.}
    \label{tab:ablation}
\end{table}
\section{Conclusion}
In this work, we propose DeLightMono, a self-supervised depth estimation algorithm that addresses uneven illumination in endoscopic scenarios through image decoupling. 
The method decomposes an endoscopic image with the illumination-reflectance-depth modeling, and harnesses these components to strengthen the depth net against uneven illumination via a joint optimization framework.
The results of comprehensive comparison experiments and ablation studies demonstrate the advancing performance of the proposed method.

\section*{Acknowledgements}
This work was supported in part by the National Natural Science Foundation of China (62401246), and Shenzhen Science and Technology Program (JCYJ20250604185805008, JCYJ20240813095112017)
\bibliography{ref}

@inproceedings{lu2023autonomous,
  title={Autonomous intelligent navigation for flexible endoscopy using monocular depth guidance and 3-D shape planning},
  author={Lu, Yiang and Wei, Ruofeng and Li, Bin and Chen, Wei and Zhou, Jianshu and Dou, Qi and Sun, Dong and Liu, Yun-hui},
  booktitle={2023 IEEE international conference on robotics and automation (ICRA)},
  pages={1--7},
  year={2023},
  organization={IEEE}
}

@article{yue2023deep,
  title={Deep pyramid network for low-light endoscopic image enhancement},
  author={Yue, Guanghui and Gao, Jie and Cong, Runmin and Zhou, Tianwei and Li, Leida and Wang, Tianfu},
  journal={IEEE Transactions on Circuits and Systems for Video Technology},
  volume={34},
  number={5},
  pages={3834--3845},
  year={2023},
  publisher={IEEE}
}

@inproceedings{engel2014lsd,
  title={LSD-SLAM: Large-scale direct monocular SLAM},
  author={Engel, Jakob and Sch{\"o}ps, Thomas and Cremers, Daniel},
  booktitle={European conference on computer vision},
  pages={834--849},
  year={2014},
  organization={Springer}
}

@article{shao2022self,
  title={Self-supervised monocular depth and ego-motion estimation in endoscopy: appearance flow to the rescue},
  author={Shao, Shuwei and Pei, Zhongcai and Chen, Weihai and Zhu, Wentao and Wu, Xingming and Sun, Dianmin and Zhang, Baochang},
  journal={Medical image analysis},
  volume={77},
  pages={102338},
  year={2022},
  publisher={Elsevier}
}

@article{metzger2024augmented,
  title={Augmented reality navigation systems in endoscopy},
  author={Metzger, Rebecca and Suppa, Per and Li, Zhen and Vemuri, Anant},
  journal={Frontiers in Gastroenterology},
  volume={3},
  pages={1345466},
  year={2024},
  publisher={Frontiers Media SA}
}

@inproceedings{cui2024endodac,
  title={Endodac: Efficient adapting foundation model for self-supervised depth estimation from any endoscopic camera},
  author={Cui, Beilei and Islam, Mobarakol and Bai, Long and Wang, An and Ren, Hongliang},
  booktitle={27th MICCAI},
  pages={208--218},
  year={2024},
  organization={Springer}
}

@inproceedings{li2022geometric,
  title={Geometric constraints for self-supervised monocular depth estimation on laparoscopic images with dual-task consistency},
  author={Li, Wenda and Hayashi, Yuichiro and Oda, Masahiro and Kitasaka, Takayuki and Misawa, Kazunari and Mori, Kensaku},
  booktitle={25th MICCAI},
  pages={467--477},
  year={2022},
  organization={Springer}
}

@article{li2024image,
  title={Image Intrinsic-Based Unsupervised Monocular Depth Estimation in Endoscopy},
  author={Li, Bojian and Liu, Bo and Zhu, Miao and Luo, Xiaoyan and Zhou, Fugen},
  journal={IEEE Journal of Biomedical and Health Informatics},
  year={2024},
  publisher={IEEE}
}

@inproceedings{godard2019digging,
  title={Digging into self-supervised monocular depth estimation},
  author={Godard, Cl{\'e}ment and Mac Aodha, Oisin and Firman, Michael and Brostow, Gabriel J},
  booktitle={Proceedings of the IEEE/CVF international conference on computer vision},
  pages={3828--3838},
  year={2019}
}

@article{recasens2021endo,
  title={Endo-depth-and-motion: Reconstruction and tracking in endoscopic videos using depth networks and photometric constraints},
  author={Recasens, David and Lamarca, Jos{\'e} and F{\'a}cil, Jos{\'e} M and Montiel, JMM and Civera, Javier},
  journal={IEEE Robotics and Automation Letters},
  volume={6},
  number={4},
  pages={7225--7232},
  year={2021},
  publisher={IEEE}
}

@article{allan2021stereo,
  author       = {Max Allan and
                  A. Jonathan McLeod and
                  Congcong Wang and
                  Jean{-}Claude Rosenthal and
                  Zhenglei Hu and
                  Niklas Gard and
                  Peter Eisert and
                  Ke Xue Fu and
                  Trevor Zeffiro and
                  Wenyao Xia and
                  Zhanshi Zhu and
                  Huoling Luo and
                  Fucang Jia and
                  Xiran Zhang and
                  Xiaohong Li and
                  Lalith Sharan and
                  Thomas Kurmann and
                  Sebastian Schmid and
                  Raphael Sznitman and
                  Dimitris Psychogyios and
                  Mahdi Azizian and
                  Danail Stoyanov and
                  Lena Maier{-}Hein and
                  Stefanie Speidel},
  title        = {Stereo Correspondence and Reconstruction of Endoscopic Data Challenge},
  journal      = {CoRR},
  volume       = {abs/2101.01133},
  year         = {2021},
  url          = {https://arxiv.org/abs/2101.01133},
  eprinttype    = {arXiv},
  eprint       = {2101.01133},
  bibsource    = {dblp computer science bibliography, https://dblp.org}
}

@inproceedings{yang2024depth,
  title={Depth anything: Unleashing the power of large-scale unlabeled data},
  author={Yang, Lihe and Kang, Bingyi and Huang, Zilong and Xu, Xiaogang and Feng, Jiashi and Zhao, Hengshuang},
  booktitle={Proceedings of the IEEE/CVF Conference on Computer Vision and Pattern Recognition},
  pages={10371--10381},
  year={2024}
}

@inproceedings{huang2022self,
  title={Self-supervised Depth Estimation in Laparoscopic Image Using 3D Geometric Consistency},
  author={Huang, Baoru and Zheng, Jian-Qing and Nguyen, Anh and Xu, Chi and Gkouzionis, Ioannis and Vyas, Kunal and Tuch, David and Giannarou, Stamatia and Elson, Daniel S},
  booktitle={25th MICCAI},
  pages={13--22},
  year={2022},
  organization={Springer}
}

@inproceedings{yue2023tcl,
  title={Tcl: Triplet consistent learning for odometry estimation of monocular endoscope},
  author={Yue, Hao and Gu, Yun},
  booktitle={26th MICCAI},
  pages={144--153},
  year={2023},
  organization={Springer}
}

@article{zhu2021intelligent,
  title={Intelligent soft surgical robots for next-generation minimally invasive surgery},
  author={Zhu, Jiaqi and Lyu, Liangxiong and Xu, Yi and Liang, Huageng and Zhang, Xiaoping and Ding, Han and Wu, Zhigang},
  journal={Advanced Intelligent Systems},
  volume={3},
  number={5},
  pages={2100011},
  year={2021},
  publisher={Wiley Online Library}
}

@inproceedings{guo2020zero,
  title={Zero-reference deep curve estimation for low-light image enhancement},
  author={Guo, Chunle and Li, Chongyi and Guo, Jichang and Loy, Chen Change and Hou, Junhui and Kwong, Sam and Cong, Runmin},
  booktitle={Proceedings of the IEEE/CVF conference on computer vision and pattern recognition},
  pages={1780--1789},
  year={2020}
}

@article{reza2004realization,
  title={Realization of the contrast limited adaptive histogram equalization (CLAHE) for real-time image enhancement},
  author={Reza, Ali M},
  journal={Journal of VLSI signal processing systems for signal, image and video technology},
  volume={38},
  pages={35--44},
  year={2004},
  publisher={Springer}
}

@inproceedings{ma2022toward,
  title={Toward fast, flexible, and robust low-light image enhancement},
  author={Ma, Long and Ma, Tengyu and Liu, Risheng and Fan, Xin and Luo, Zhongxuan},
  booktitle={Proceedings of the IEEE/CVF conference on computer vision and pattern recognition},
  pages={5637--5646},
  year={2022}
}

@inproceedings{poggi2020uncertainty,
  title={On the uncertainty of self-supervised monocular depth estimation},
  author={Poggi, Matteo and Aleotti, Filippo and Tosi, Fabio and Mattoccia, Stefano},
  booktitle={CVPR 2020},
  pages={3227--3237},
  year={2020}
}

@inproceedings{lyu2021hr,
  title={Hr-depth: High resolution self-supervised monocular depth estimation},
  author={Lyu, Xiaoyang and Liu, Liang and Wang, Mengmeng and Kong, Xin and Liu, Lina and Liu, Yong and Chen, Xinxin and Yuan, Yi},
  booktitle={Proceedings of the AAAI conference on artificial intelligence},
  volume={35},
  number={3},
  pages={2294--2301},
  year={2021}
}

@inproceedings{amitai2023self,
  title={Self-supervised monocular depth underwater},
  author={Amitai, Shlomi and Klein, Itzik and Treibitz, Tali},
  booktitle={2023 IEEE International Conference on Robotics and Automation (ICRA)},
  pages={1098--1104},
  year={2023},
  organization={IEEE}
}

@inproceedings{wu2023learning,
  title={Learning semantic-aware knowledge guidance for low-light image enhancement},
  author={Wu, Yuhui and Pan, Chen and Wang, Guoqing and Yang, Yang and Wei, Jiwei and Li, Chongyi and Shen, Heng Tao},
  booktitle={CVPR 2023},
  pages={1662--1671},
  year={2023}
}

@article{guo2016lime,
  title={LIME: Low-light image enhancement via illumination map estimation},
  author={Guo, Xiaojie and Li, Yu and Ling, Haibin},
  journal={IEEE Transactions on image processing},
  volume={26},
  number={2},
  pages={982--993},
  year={2016},
  publisher={IEEE}
}

@inproceedings{liu2021self,
  title={Self-supervised Monocular Depth Estimation for All Day Images using Domain Separation},
  author={Liu, Lina and Song, Xibin and Wang, Mengmeng and Liu, Yong and Zhang, Liangjun},
  booktitle={Proceedings of the IEEE/CVF International Conference on Computer Vision},
  pages={12737--12746},
  year={2021}
}

@inproceedings{zhao2022monovit,
  title={Monovit: Self-supervised monocular depth estimation with a vision transformer},
  author={Zhao, Chaoqiang and Zhang, Youmin and Poggi, Matteo and Tosi, Fabio and Guo, Xianda and Zhu, Zheng and Huang, Guan and Tang, Yang and Mattoccia, Stefano},
  booktitle={2022 3DV},
  pages={668--678},
  year={2022},
  organization={IEEE}
}

@inproceedings{kingma2020method,
  author       = {Diederik P. Kingma and
                  Jimmy Ba},
  editor       = {Yoshua Bengio and
                  Yann LeCun},
  title        = {Adam: {A} Method for Stochastic Optimization},
  booktitle    = {3rd International Conference on Learning Representations, {ICLR} 2015,
                  San Diego, CA, USA, May 7-9, 2015, Conference Track Proceedings},
  year         = {2015},
  url          = {http://arxiv.org/abs/1412.6980},
  bibsource    = {dblp computer science bibliography, https://dblp.org}
}

@inproceedings{Chen2018Retinex,
 title={Deep Retinex Decomposition for Low-Light Enhancement},
 author={Chen, Wei and Wenjing, Wang and Wenhan, Yang and Jiaying, Liu},
 booktitle={British Machine Vision Conference},
 year={2018},
 organization={British Machine Vision Association}
}

@inproceedings{zhou2017unsupervised,
  title={Unsupervised learning of depth and ego-motion from video},
  author={Zhou, Tinghui and Brown, Matthew and Snavely, Noah and Lowe, David G},
  booktitle={CVPR 2017},
  pages={1851--1858},
  year={2017}
}

@article{ozyoruk2021endoslam,
  title={EndoSLAM dataset and an unsupervised monocular visual odometry and depth estimation approach for endoscopic videos},
  author={Ozyoruk, Kutsev Bengisu and Gokceler, Guliz Irem and Bobrow, Taylor L and Coskun, Gulfize and Incetan, Kagan and Almalioglu, Yasin and Mahmood, Faisal and Curto, Eva and Perdigoto, Luis and Oliveira, Marina and others},
  journal={Medical image analysis},
  volume={71},
  pages={102058},
  year={2021},
  publisher={Elsevier}
}

@article{he2024monolot,
  title={MonoLoT: Self-Supervised Monocular Depth Estimation in Low-Texture Scenes for Automatic Robotic Endoscopy},
  author={He, Qi and Feng, Guang and Bano, Sophia and Stoyanov, Danail and Zuo, Siyang},
  journal={IEEE Journal of Biomedical and Health Informatics},
  year={2024},
  publisher={IEEE}
}

@article{ranftl2020towards,
  title={Towards robust monocular depth estimation: Mixing datasets for zero-shot cross-dataset transfer},
  author={Ranftl, Ren{\'e} and Lasinger, Katrin and Hafner, David and Schindler, Konrad and Koltun, Vladlen},
  journal={IEEE transactions on pattern analysis and machine intelligence},
  volume={44},
  number={3},
  pages={1623--1637},
  year={2020},
  publisher={IEEE}
}

@inproceedings{budd2024transferring,
  title={Transferring Relative Monocular Depth to Surgical Vision with Temporal Consistency},
  author={Budd, Charlie and Vercauteren, Tom},
  booktitle={27th MICCAI},
  pages={692--702},
  year={2024},
  organization={Springer}
}

@article{eigen2014depth,
  title={Depth map prediction from a single image using a multi-scale deep network},
  author={Eigen, David and Puhrsch, Christian and Fergus, Rob},
  journal={Advances in neural information processing systems},
  volume={27},
  year={2014}
}

@inproceedings{li2015depth,
  title={Depth and surface normal estimation from monocular images using regression on deep features and hierarchical crfs},
  author={Li, Bo and Shen, Chunhua and Dai, Yuchao and Van Den Hengel, Anton and He, Mingyi},
  booktitle={CVPR 2015},
  pages={1119--1127},
  year={2015}
}

@article{karsch2014depth,
  title={Depth transfer: Depth extraction from video using non-parametric sampling},
  author={Karsch, Kevin and Liu, Ce and Kang, Sing Bing},
  journal={IEEE transactions on pattern analysis and machine intelligence},
  volume={36},
  number={11},
  pages={2144--2158},
  year={2014},
  publisher={IEEE}
}

@article{pizer1987adaptive,
  title={Adaptive histogram equalization and its variations},
  author={Pizer, Stephen M and Amburn, E Philip and Austin, John D and Cromartie, Robert and Geselowitz, Ari and Greer, Trey and ter Haar Romeny, Bart and Zimmerman, John B and Zuiderveld, Karel},
  journal={Computer vision, graphics, and image processing},
  volume={39},
  number={3},
  pages={355--368},
  year={1987},
  publisher={Elsevier}
}

@article{zhang2021beyond,
  title={Beyond brightening low-light images},
  author={Zhang, Yonghua and Guo, Xiaojie and Ma, Jiayi and Liu, Wei and Zhang, Jiawan},
  journal={International Journal of Computer Vision},
  volume={129},
  pages={1013--1037},
  year={2021},
  publisher={Springer}
}

@inproceedings{cai2023retinexformer,
  title={Retinexformer: One-stage retinex-based transformer for low-light image enhancement},
  author={Cai, Yuanhao and Bian, Hao and Lin, Jing and Wang, Haoqian and Timofte, Radu and Zhang, Yulun},
  booktitle={Proceedings of the IEEE/CVF international conference on computer vision},
  pages={12504--12513},
  year={2023}
}

@inproceedings{enqvist2011non,
  title={Non-sequential structure from motion},
  author={Enqvist, Olof and Kahl, Fredrik and Olsson, Carl},
  booktitle={2011 ICCV Workshops},
  pages={264--271},
  year={2011},
  organization={IEEE}
}

@inproceedings{fu2018deep,
  title={Deep ordinal regression network for monocular depth estimation},
  author={Fu, Huan and Gong, Mingming and Wang, Chaohui and Batmanghelich, Kayhan and Tao, Dacheng},
  booktitle={CVPR 2018},
  pages={2002--2011},
  year={2018}
}

@InProceedings{pmlr-v205-vankadari23a,
  title = 	 {When the Sun Goes Down: Repairing Photometric Losses for All-Day Depth Estimation},
  author =       {Vankadari, Madhu and Golodetz, Stuart and Garg, Sourav and Shin, Sangyun and Markham, Andrew and Trigoni, Niki},
  booktitle = 	 {Proceedings of The 6th CoRL},
  pages = 	 {1992--2003},
  year = 	 {2023},
  editor = 	 {Liu, Karen and Kulic, Dana and Ichnowski, Jeff},
  volume = 	 {205},
  series = 	 {Proceedings of Machine Learning Research},
  month = 	 {14--18 Dec},
  publisher =    {PMLR},
  url = 	 {https://proceedings.mlr.press/v205/vankadari23a.html}
}

@inproceedings{zheng2023steps,
  title={Steps: Joint self-supervised nighttime image enhancement and depth estimation},
  author={Zheng, Yupeng and Zhong, Chengliang and Li, Pengfei and Gao, Huan-ang and Zheng, Yuhang and Jin, Bu and Wang, Ling and Zhao, Hao and Zhou, Guyue and Zhang, Qichao and others},
  booktitle={2023 IEEE International Conference on Robotics and Automation (ICRA)},
  pages={4916--4923},
  year={2023},
  organization={IEEE}
}

@inproceedings{godard2017unsupervised,
  title={Unsupervised monocular depth estimation with left-right consistency},
  author={Godard, Cl{\'e}ment and Mac Aodha, Oisin and Brostow, Gabriel J},
  booktitle={CVPR 2017},
  pages={270--279},
  year={2017}
}

@article{land1971lightness,
  title={Lightness and retinex theory},
  author={Land, Edwin H and McCann, John J},
  journal={Journal of the Optical society of America},
  volume={61},
  number={1},
  pages={1--11},
  year={1971},
  publisher={Optical Society of America}
}

@article{land1977retinex,
  title={The retinex theory of color vision},
  author={Land, Edwin H},
  journal={Scientific american},
  volume={237},
  number={6},
  pages={108--129},
  year={1977},
  publisher={JSTOR}
}

@article{oren1995generalization,
  title={Generalization of the Lambertian model and implications for machine vision},
  author={Oren, Michael and Nayar, Shree K},
  journal={International Journal of Computer Vision},
  volume={14},
  number={3},
  pages={227--251},
  year={1995},
  publisher={Springer}
}

@inproceedings{wang2021regularizing,
  title={Regularizing nighttime weirdness: Efficient self-supervised monocular depth estimation in the dark},
  author={Wang, Kun and Zhang, Zhenyu and Yan, Zhiqiang and Li, Xiang and Xu, Baobei and Li, Jun and Yang, Jian},
  booktitle={Proceedings of the IEEE/CVF international conference on computer vision},
  pages={16055--16064},
  year={2021}
}

@inproceedings{klingner2020self,
  title={Self-supervised monocular depth estimation: Solving the dynamic object problem by semantic guidance},
  author={Klingner, Marvin and Term{\"o}hlen, Jan-Aike and Mikolajczyk, Jonas and Fingscheidt, Tim},
  booktitle={European conference on computer vision},
  pages={582--600},
  year={2020},
  organization={Springer}
}

@article{zhou2025manydepth2,
  title={Manydepth2: Motion-aware self-supervised monocular depth estimation in dynamic scenes},
  author={Zhou, Kaichen and Bian, Jia-Wang and Zheng, Jian-Qing and Zhong, Jiaxing and Xie, Qian and Trigoni, Niki and Markham, Andrew},
  journal={IEEE Robotics and Automation Letters},
  year={2025},
  publisher={IEEE}
}

@article{fan2024triple,
  title={Triple-Supervised Convolutional Transformer Aggregation for Robust Monocular Endoscopic Dense Depth Estimation},
  author={Fan, Wenkang and Jiang, Wenjing and Shi, Hong and Zeng, Hui-Qing and Chen, Yinran and Luo, Xiongbiao},
  journal={IEEE Transactions on Medical Robotics and Bionics},
  volume={6},
  number={3},
  pages={1017--1029},
  year={2024},
  publisher={IEEE}
}

\end{document}